# The effect of different feature selection methods on models created with XGBoost

Jorge Neyra[1], Vishal B. Siramshetty[2], and Huthaifa I. Ashqar[1,3]

[1] Data Science Department, University of Maryland Baltimore County

[2] Genentech, Inc., USA

[3] Arab American University, Jenin, Palestine

# Abstract

This study examines the effect that different feature selection methods have on models created with XGBoost, a popular machine learning algorithm with superb regularization methods. It shows that three different ways for reducing the dimensionality of features produces no statistically significant change in the prediction accuracy of the model. This suggests that the traditional idea of removing the noisy training data to make sure models do not overfit may not apply to XGBoost. But it may still be viable in order to reduce computational complexity.

# Introduction

XGBoost is a gradient tree boosting machine learning method with increased computational performance and superior regularization methods (to control overfitting). It has become a popular algorithm and software package[1] that has been proven efficient in many use cases and in many fields. And it's also the method of choice in many competitions [1].

In more traditional machine learning and data mining methods, overfitting is a problem if one doesn't reduce the dimensionality of the training data ("the curse of dimensionality") [2]. This means that if you train your models with redundant features (noise), the final model may not be able to extrapolate (make inferences about data it has not seen before). This study answers the question if this is a problem in XGBoost.

---

[1] https://github.com/dmlc/xgboost

One thing to consider is that when creating machine learning models, there are two goals that one can pursue: accuracy and/or interpretability. When you model for accuracy, you don't concern yourself with the relationship between the independent variables (features) and the dependent variables (labels) [7-10]. Conversely, when you have a goal of interpretability, you really want to know which are the most significant features and how they relate to your model's results [3].

This study only considers the accuracy of the XGBoost models and how it is affected by the selection of features.

# Methods

In order to make this analysis, I had to select data with high dimensionality (large number of features), and potentially multiple labels so I could create multiple models at once (one for each label). Also, I had to choose the right feature selection methods and the right statistical methods to measure the significance of the results.

## Datasets

I used the data from the Tox21 Data Challenge in 2014[2], which is a chemical compound library of around 10,000 items, along with 12 observed binary values (labels). Each of these labels indicates if the associated compound is interfering with a specific biochemical pathway. Labels contain a 1 if the compound is interfering (which means it is very likely to be toxic) and 0 if it's not.

For the features, I used Dragon descriptors[3], a list of up to 4885 numerical values that describe any given chemical compound.

---

[2] https://tripod.nih.gov/tox21/challenge/data.jsp.
[3] http://www.talete.mi.it/products/dragon_molecular_descriptor_list.pdf.

In order to prepare the data so it could be used to create the models, the data had to be split into different data sets - one corresponding to each label. And for each subsequent dataset, I removed all compounds (along with its descriptors) that did not have a label. Then, each dataset was split in 3 parts: 80% to train models, 10% to validate the model (making sure it makes good predictions on data it has seen), and 10% to test the best performing model. Table 1 shows the number of compounds in each split for each dataset.

## Feature selection

Three different methods were used for feature selection, based on their success on other studies: random forest classifier method [4, 10s], ANOVA method [5], and Chi-Square method [6]. The random forest classifier method was used first in every dataset because it produces a selected number of features that are deemed significant. The other two classifiers rank the features based on their level of significance, and then one can select an arbitrary number of features. For this experiment, I used the same quantity of features selected by the random forest method. But keep in mind that not the same features were used - although there is overlap.

Table 1 shows the total number of features available for each dataset and the number of features selected by each method. On average, the number of features were reduced by about 64%.

Table 1. This table shows the description of each of the datasets used for the analysis. The number of compounds has the splits used to train and validate the models and number of

| Dataset | Number of Compounds | | | Number of Features | | | |
|---|---|---|---|---|---|---|---|
| | Train | Validation | Test | Total | Random Forest | ANOVA | chi square |
| NR-AR | 5812 | 726 | 727 | 3150 | 891 | 891 | 891 |
| NR-AR-LBD | 5406 | 676 | 676 | 3143 | 887 | 887 | 887 |
| NR-AhR | 5239 | 655 | 655 | 3122 | 1066 | 1066 | 1066 |
| NR-Aroma | 4656 | 582 | 583 | 3103 | 1201 | 1201 | 1201 |
| NR-ER | 4954 | 619 | 620 | 3093 | 1240 | 1240 | 1240 |
| NR-ER-LBD | 5564 | 695 | 696 | 3128 | 1112 | 1112 | 1112 |
| NR-PPAR-g | 5160 | 645 | 645 | 3103 | 1145 | 1145 | 1145 |
| SR-ARE | 4665 | 583 | 584 | 3104 | 1187 | 1187 | 1187 |
| SR-ATAD5 | 5657 | 707 | 708 | 3138 | 1220 | 1220 | 1220 |
| SR-HSE | 5173 | 647 | 647 | 3117 | 1184 | 1184 | 1184 |
| SR-MMP | 4648 | 581 | 581 | 3114 | 1053 | 1053 | 1053 |
| SR-p53 | 5419 | 677 | 678 | 3132 | 1183 | 1183 | 1183 |

## Model creation

20 models were created for each feature selection for each dataset (a total of 80 models per dataset). And 20 different sets of randomly generated hyperparameters were used to fit the models. This means that for any given dataset, a set of hyperparameters was generated, 4 models were trained (representing each feature selection method), and then the process was repeated 20 times.

For each model, the 80% split was used to train it, and the 10% validation split was used to measure its accuracy on new data. I used the receiver operating characteristic (ROC) area under the curve (AUC) score as the unit of accuracy, which is the standard for measuring binary classification problems.

Based on this, 260 scores were generated for each feature selection method. And these scores were used to measure the statistical significance of each feature selection.

## Descriptive statistics

Independent t-tests were used to measure the statistical significance of the change in the scores for each feature selection method. This method was chosen because the standard deviation of the population that was sampled is not known, and because the number of samples is greater than 30. Each set of scores passed the assumption of normality based on the Central Limit Theorem (> 200 samples). Also, the Pearson correlation coefficient was calculated to measure the level of correlation between the scores obtained without feature selection and the scores obtained with the different feature selection methods.

And finally, Cohen's d was calculated to determine if the effect size of each feature selection method was statistically significant.

## Results and Analysis

The best models for each dataset and each feature selection method were validated against the test datasets. Then these scores were compared against the submissions to the Tox21 Data Challenge. This provided a good idea on the level of performance of each model.

And when observing the comparison (Figure 1), one can see the superior performance by most of our models (labeled "my_scores"), as they obtained higher scores on at least half of the labels/datasets. And one can also observe the similarity among our them, with their corresponding lines following almost identical patterns. And since the scores were calculated on data that the models had not seen before, it is easy to assume that the neither model was overfitting more than the other.

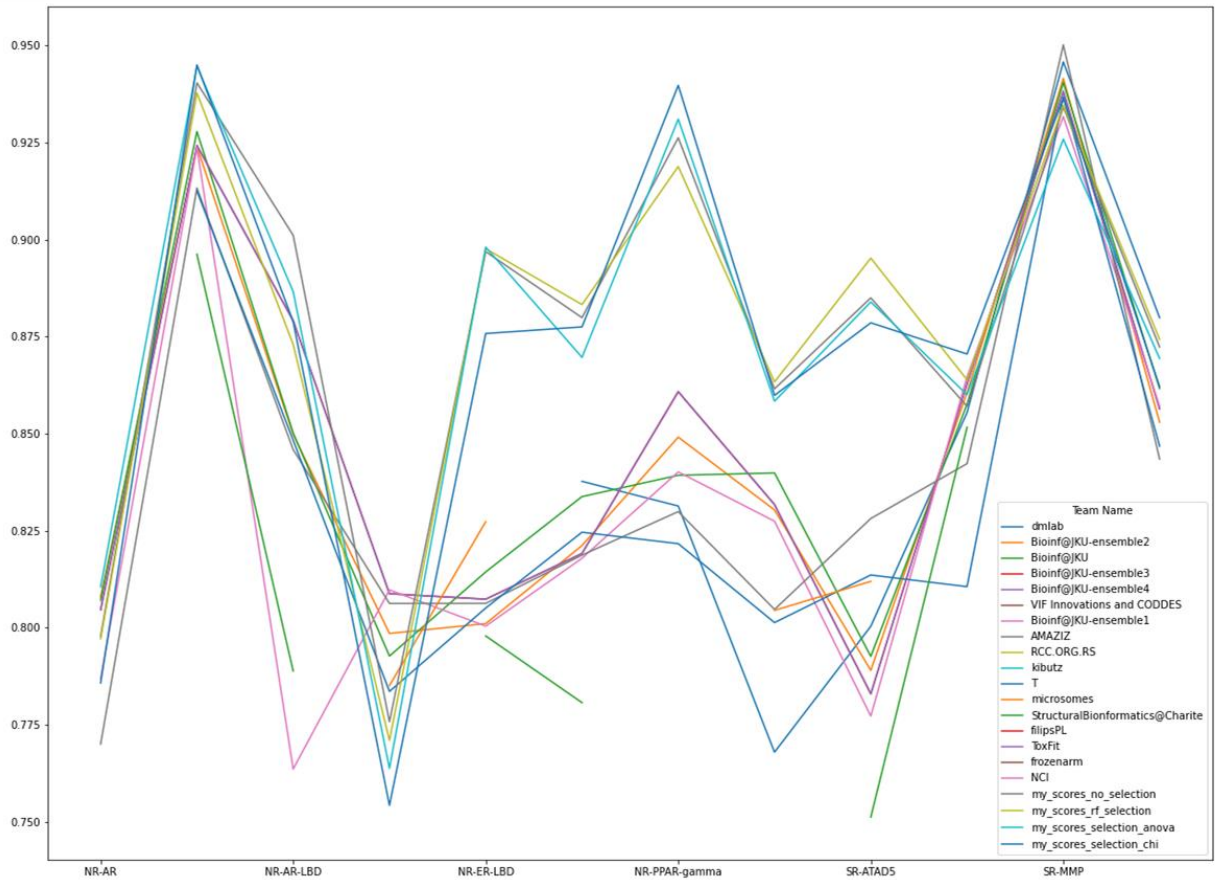

Figure 1. Comparison between the accuracy scores of the best models created in this study (my_scores) and the scores of the teams that participated in the Tox21 Challenge.

To further analyze the similarities of the models, a comparison of the distributions of the different sets of scores was performed (Figure 2), and one can quickly notice their uniformity. All distributions are slightly negatively skewed and slightly leptokurtic (lean and tall). This is a very clear visual indication that the different types of feature selection methods have very little effect on the models created by the XGBoost algorithm.

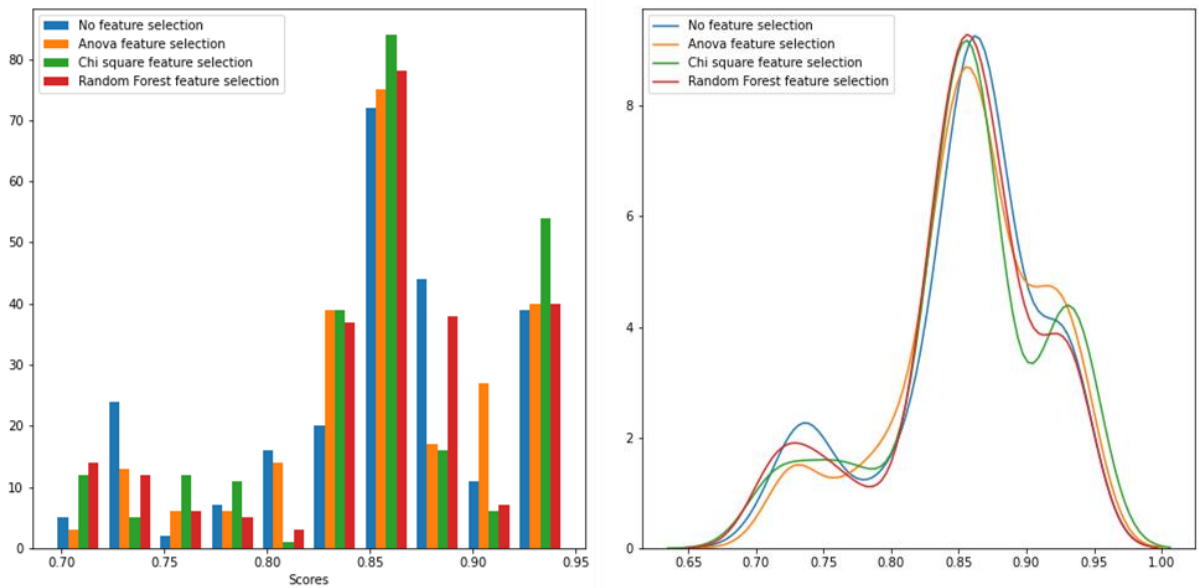

Figure 2. A histogram (A) and a kernel density estimation plot (B) comparing the

Also, a correlation analysis between the set of scores from models without feature selection against all the other sets of scores shows high correlation among them (Figure 3). The Pearson correlation coefficient (r) for each is 0.94 or above (on a scale from 0 to 1, with 1 being the highest correlation). Furthermore, a visual inspection of the corresponding scatter plots indicates that all datasets have homogeneity of variance with each other, further driving the point of the little statistical difference between the different sets.

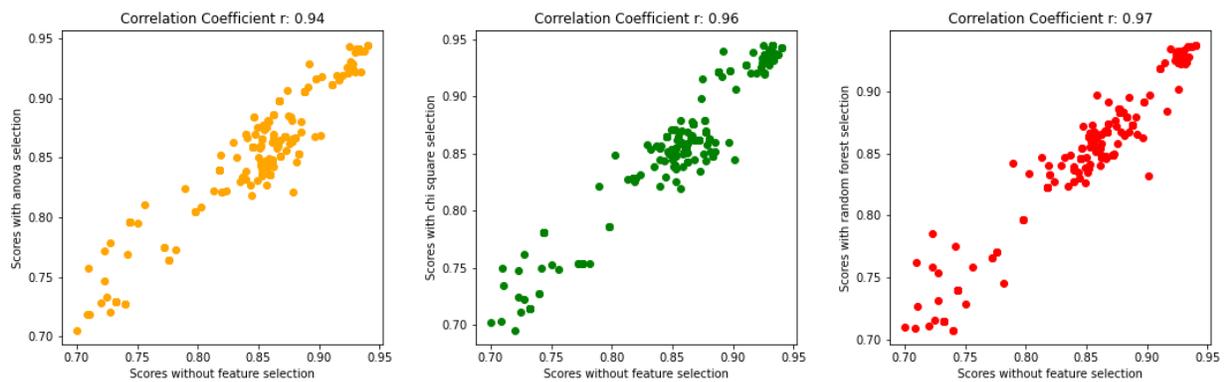

Figure 3. Correlation plots showing the high level of correlation between the scores developed without any feature selection and scores developed

In order to quantify and verify the statistical significance and the effect size of applying the different feature selection methods to the XGBoost models, I performed t-tests and calculated Cohen's d (the significance of the effect size). The results can be seen in Table 2 and they show that there is no significant difference between the mean of the scores from the models developed without feature selection and the others. Setting the significance level at 0.95, all p-values fail to meet the threshold of significance (< 0.05).

And the small difference that exists between the means (the effect size) also shows no statistical significance as all d values are below 0.2.

Table 2. Results from independent t-test and effect size (Cohen's d), comparing scores from models developed without any feature selection (mean = 0.853) with all other sets of scores.

| Feature selection method compared against no feature selection | Mean | t-test score | Degrees of freedom | p-value | d |
|---|---|---|---|---|---|
| ANOVA | 0.855 | 0.9 | 239 | 0.39 | 0.08 |
| Chi-square | 0.854 | 0.34 | 239 | 0.73 | 0.03 |
| Random forest | 0.85 | 0.9 | 239 | 0.37 | 0.04 |

# Conclusion

As noted above, XGBoost has become one of the most widely used machine learning methods. Compared to other tree boosting methods, it has very efficient regularization methods that helps with overfitting. So in order to test this efficiency, I picked a dataset with multiple labels and a high number of features. This data was split into multiple datasets and then used to create a total of 960 models, with 3 different methods for feature selection and 1 without any feature selection.

A number of statistical methods were used to measure the significance of the changes in scores obtained by the models created with the different feature selection methods. All measurements showed that the changes in the means of the scores were not statistically significant.

This suggests that the "curse of dimensionality" does not apply when creating a model using the XGBoost algorithm (at least at the dimensionality levels indicated above). Removing redundant or insignificant features from the dataset did not increase my models' abilities to extrapolate on new data. This shows that the regularization functions in the XGBoost algorithm are very efficient at neutralizing the noise from the data.

But these results also show that it is safe, and probably preferred, to remove the noisy features from the data without affecting the performance of the model. This helps with computational complexity and time when training a model. And it's specially important as the number of samples and/or the number of features increases to the tens or hundreds of thousands (or millions).

For the future, a different study can test the performance of additional methods for feature selection and maybe the minimum number of features that can be used to still have a good performing model. Also, this test was only performed on binary classification tasks, so other tasks like regression and multiclass classification also need to be tested.